\documentclass[runningheads]{llncs}
\usepackage{graphicx}

\usepackage{tikz}
\usepackage{comment}
\usepackage{amsmath,amssymb} 
\usepackage{color}
\usepackage{orcidlink}
\usepackage{ltablex,  booktabs, ragged2e}
\usepackage[accsupp]{axessibility}  

\begin{document}
\pagestyle{headings}
\mainmatter

\title{Real-time Driver Monitoring Systems on Edge AI Device } 

\titlerunning{Real-time DMS}
%
\author{Jyothi Hariharan\index{Hariharan, Jyothi} \and
Rahul Rama Varior \and
Sunil Karunakaran }
\authorrunning{J. Hariharan et al.}
%
\institute{
Ignitarium Technology Solutions Private Limited, Bangalore, India\\
\email{\{jyothi.hariharan, rahul.varior, sunil.karunakaran\}@ignitarium.com}}

\maketitle

\begin{abstract}
As road accident cases are increasing due to the inattention of the driver, automated driver monitoring systems (DMS) have gained an increase in acceptance. In this report, we present a real-time DMS system that runs on a hardware-accelerator-based edge device.  The system consists of an InfraRed camera to record the driver footage and an edge device to process the data.  To successfully port the deep learning models to run on the edge device taking full advantage of the hardware accelerators, model surgery was performed. The final DMS system achieves 63 frames per second (FPS) on the TI-TDA4VM edge device. 
\end{abstract}

\section{Introduction}

Road accidents are increasing due to the inattention of the driver. Around the world, there is an increase in the acceptance of automated systems to alert the driver in the event of a distraction from the driving task. For instance, Euro New Car Assessment Programme (NCAP) standards and the European Commission (EC) regulation mandates Driver Monitoring Systems (DMS) technology from 2023. The U.S. National Transportation Safety Board (NTSB) recommends using DMS in semi-autonomous cars. DMS built using Artificial intelligence (AI) technology is proving to be an effective safety feature and has gained much attention in the automotive industry in recent years. In this article, we present the in-house research and development efforts to build a camera-based real-time DMS system that runs on an edge device.  

Safety features in automobiles should allow sufficient response time for the driver to take necessary actions. Hence, the data must be processed, and the driver must be alerted in real time if the driver is distracted or drowsy. While numerous topologies (including CPU-only, CPU + GPU, CPU + Accelerator etc.) exist, the specific compute-architecture selected will usually have a direct bearing on the cost and power consumption of the chosen edge device. Deploying Deep Learning (DL) based AI systems on low-cost edge devices can be challenging due to memory limitations and low processing capacity. Moreover, porting the Deep Learning models to run on the edge device to take full advantage of the specialized AI hardware accelerators can be challenging due to a host of reasons. All required DL operators and parameter ranges maybe not be supported by the vendor’s Software Development Kit (SDK). In many cases, support may be available in the SDK but the implementation may not be highly optimized for the underlying hardware.  

In this article, we present our R\&D efforts to build a camera based real-time DMS system that runs on a hardware-accelerator-based edge device. First, the high-level architecture of the DMS is presented along with the details of the individual components of the system. The challenges and solutions to successfully port the model to run on an exemplary edge device hardware accelerator (viz. Texas Instruments TDA4VM) are explained in the next section. Finally, we present the performance improvements in terms of inference time and frames per second (FPS) that we achieved after successfully porting the model to run on the hardware accelerator. 

\section{High-level architecture of DMS}
\begin{figure}[ht]
    \centering
    \includegraphics[width=1\textwidth]{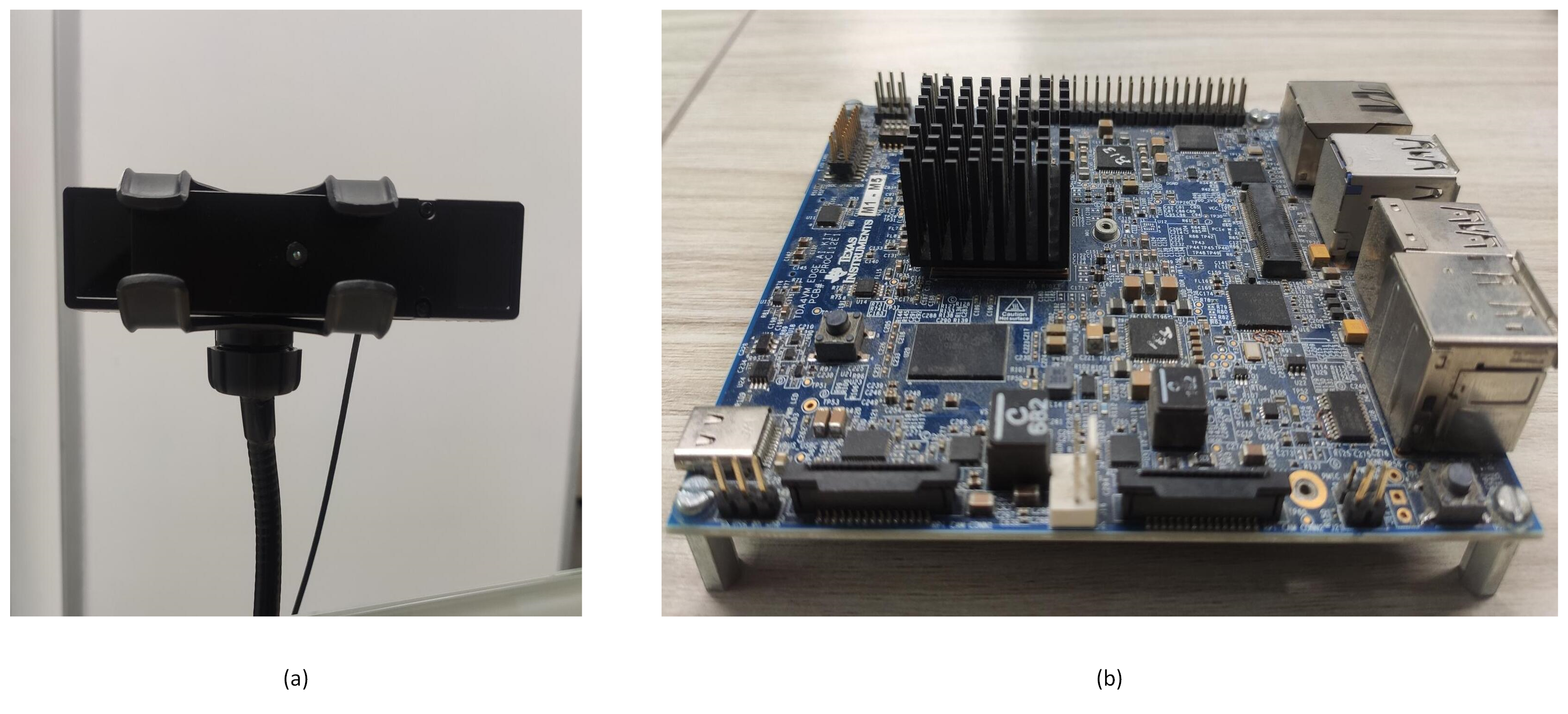}
    \caption{Components of DMS (a) The \href{https://www.leopardimaging.com/product/autonomous-camera/maxim-gmsl2-cameras/li-ar0144ivec-gmsl2/li-usb30-ar0144ivec-gmsl2-055h/}{LI-USB30-AR0144IVEC-GMSL2-055H IR camera} (b) The \href{https://www.ti.com/product/TDA4VM}{TI TDA4VM} edge AI dev kit}
    \label{fig:components}
\end{figure}

The DMS consists of an infrared camera from Leopard Imaging, that is used to capture video data. The collected data is processed by the AI stack which runs on the edge AI device. The results, for the purpose of demonstration, are displayed on a standard display device. Note that the display device is only for visualization purposes and is not part of the end system. Fig \ref{fig:components} shows the components of the DMS. Subsequent sections describe each of the components in detail.

\subsection{Components of DMS }

\subsubsection{Infrared camera }

To capture driver footage, an infrared (IR) camera \href{https://www.leopardimaging.com/product/autonomous-camera/maxim-gmsl2-cameras/li-ar0144ivec-gmsl2/li-usb30-ar0144ivec-gmsl2-055h/}{LI-USB30-AR0144IVEC-GMSL2-055H}
from \href{https://www.leopardimaging.com/}{Leopard Imaging} (LI) is mounted on the dashboard inside the driver cabin. The advantages of the LI IR camera are twofold. First, the camera module has IR illumination which enables low-light and night vision. The IR filter in the camera module also prevents interference due to ambient lighting. Second, IR cameras enable ‘see-through’ of certain sunglasses so that the eye pupil location can be accurately captured. For more details on the camera specifications and its datasheet, interested readers can refer to the product \href{https://www.leopardimaging.com/wp-content/uploads/LI-AR0144IVEC-GMSL2-055H_Datasheet.pdf}{datasheet}.  

\subsubsection{AI Stack }

The AI stack consists of the edge device and the perception system or the software application. 

\textbf{Edge device }: The edge device used is \href{https://www.ti.com/product/TDA4VM}{TDA4VM}
from \href{https://www.ti.com/}{Texas Instruments}. TDA4VM features a dual 64-bit Arm\textsuperscript{\tiny\textregistered} Cortex\textsuperscript{\tiny\textregistered}-A72 microprocessor subsystem, a C7x floating point vector DSP of up to 80 GFLOPS, and a deep-learning Matrix Multiply Accelerator (MMA) of up to 8 TOPS (8-bits). The specialized hardware accelerator MMA processes much of the deep learning workloads and is optimized for 8-bit precision (int8) and 16-bit (int16) precision integer computations. Interested readers can find more information about the product in the \href{https://www.ti.com/lit/ds/symlink/tda4vm.pdf?ts=1672632585338&ref_url=https%253A%252F%252Fwww.google.com%252F}{TDA4VM datasheet}.
We used the \href{https://www.ti.com/tool/download/PROCESSOR-SDK-LINUX-RT-J721E/08.02.00.01}{SDK version 8.02.00}.

\begin{figure}[ht]
    \centering
    \includegraphics[width=1\textwidth]{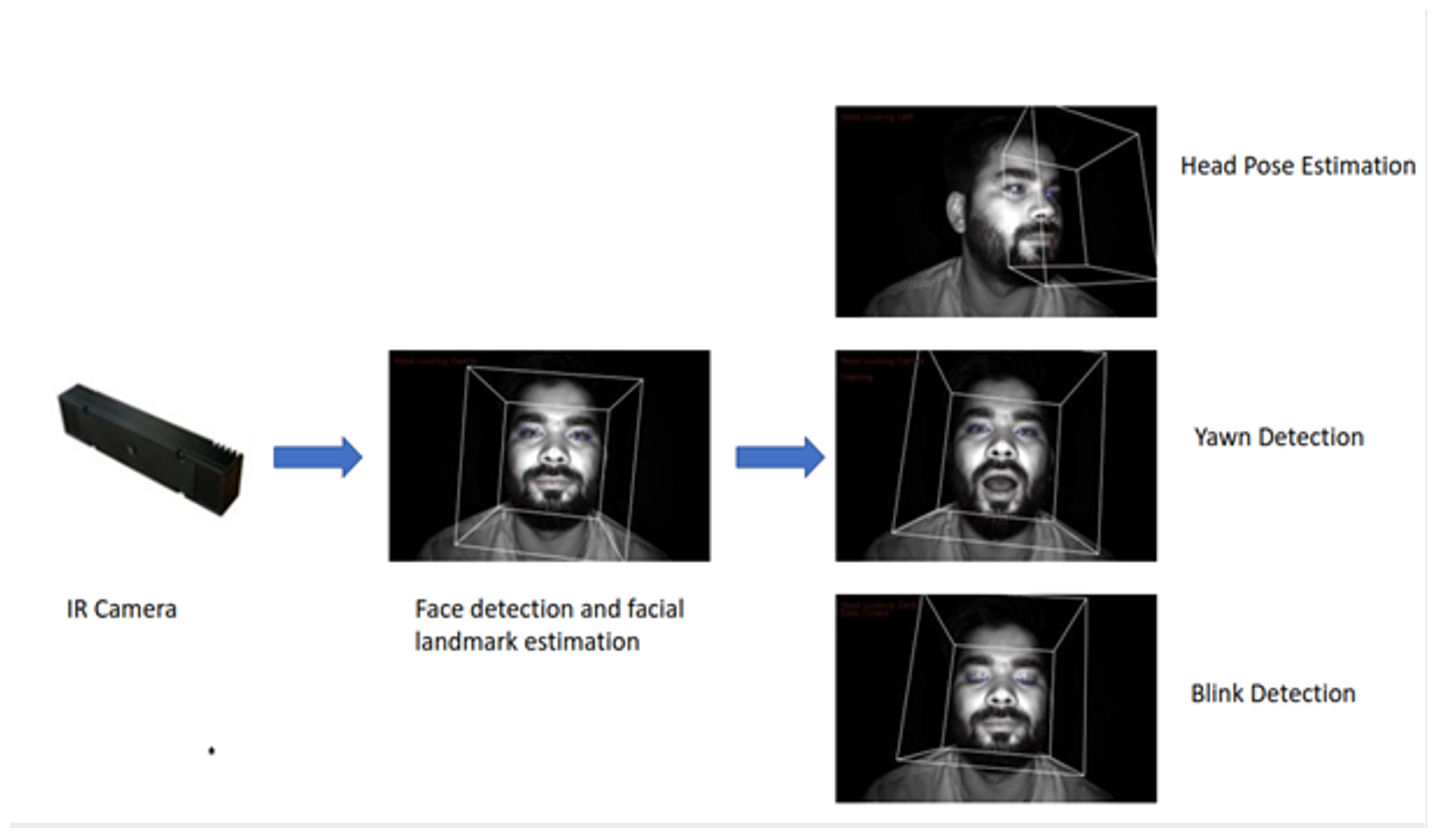}
    \caption{Pipeline Stages of the Perception System}
    \label{fig:architecture}
\end{figure}

\textbf{Perception system }: The perception system consists of a deep learning-based face detection model followed by a face landmark model. Individual frames from the video input are passed through the face detection algorithm to detect the driver’s face. The detected face is cropped and passed as input for the face landmark model. The face landmark algorithm estimates key facial points such as the face silhouette, the iris points, eye points, eyebrow points, upper and lower lips points, nose bridge, and nose bottom. These facial landmark points are used to estimate the head pose, eye closure duration, and eye gaze of each eye, and to detect whether the driver is yawning or not. The yawn results, eye closure duration, the eye gaze, and the head pose of the driver are used to infer the state of the driver, i.e., whether the driver is distracted, drowsy, or alert. The pipeline stages of the perception system are shown in Fig \ref{fig:architecture}. 

\textbf{Face detection deep learning models}: For face detection, we conducted experiments with Mediapipe face detection model BlazeFace \cite{blazeface} and YOLOX-tiny model \cite{edgeaimodelzoo} available in the TI edge AI vision model-zoo.  

\textbf{Face landmark deep learning models}: For face landmark estimation, we conducted experiments with the Google Mediapipe face mesh model \cite{mpfacemesh} and the Practical Facial Landmark Detector (PFLD) \cite{pfld} model. In the following section, the model porting and optimization efforts are explained.

\section{Model porting for TDA4VM }
The TDA4VM is equipped with the MMA which is a hardware accelerator that processes deep learning workloads. However, it is possible that not all the deep learning operators or operator configurations are supported for execution in the MMA. The unsupported operators, in such cases, will normally be executed in the ARM CPU core resulting in a reduction in the overall inference speed of the deep learning model. To overcome this, the pre-trained models will have to be adapted so that their functionality is not affected and is fully executed on the accelerator. The following section describes our efforts to port the model so that it is fully offloaded to the MMA. 

While loading the DL models to run on the MMA, the input data to the model will be copied from the CPU to the accelerator memory. In case an operator in the DL model is not supported by the device hardware accelerator library, the operation will be performed by the ARM CPU. The input data to the corresponding layer will have to be copied from the accelerator memory to the CPU memory for executing the layer. After the layer is executed, the outputs need to be copied back to the hardware accelerator memory for processing the rest of the layers. The data transfer between the CPU and the accelerator is expensive and will result in suboptimal inference speed.  

To take full advantage of the hardware accelerator, all layers of the model must be offloaded to the MMA. Therefore, unsupported layers/operators in the model’s frozen graph must be replaced/removed (model graph surgery or model surgery) with functionally accurate layers/operators that are supported in the MMA. In order to perform the model surgery, we develop a model graph translation library (or toolkit). The model graph translation library helps to create a trainable model from a frozen tflite or ONNX graph and the surgery can be performed on the trainable model. The benefits of using this library are:

  \begin{enumerate}
    \item \textbf{Conversion between different platforms} – For instance, a tflite model can be converted to an ONNX model or vice versa.  
    \item \textbf{Creating a trainable model from a frozen graph} – A tflite or ONNX frozen graph can be converted to a trainable Keras \cite{keras} model or PyTorch \cite{pytorch} model respectively. The benefit of this conversion is that in case we want to train or finetune a model after the layer update, the corresponding PyTorch or Keras model can be used. 
    \item \textbf{Extracting or modifying layers} or operators is more convenient compared to directly editing the frozen graph.
    \item  \textbf{Post-training model quantization} in TensorFlow \cite{tensorflow} expects the input model to be in Keras. The process mentioned in step 2, the creation of a trainable model from the frozen graph (Keras in this case) helps in using the post-training quantization process for TensorFlow. 
\end{enumerate}

In the next section, we present some examples of the model graph surgery we performed in order to port the aforementioned models to the edge device and achieve maximum performance in terms of speed without any change in its functionality. 

\subsection{Model graph surgery }

We conducted our model surgery experiments on the Mediapipe BlazeFace face detection model \cite{blazeface}, Mediapipe face mesh model \cite{mpfacemesh} and PFLD face landmark model \cite{pfld} 

\subsubsection{Dequantize Operators and FP16 operations in Mediapipe models }

Mediapipe models are released with float16 parameters. The face detection and face landmark models have \textbf{dequantize} operators that convert the float16 model parameters to float32 so that the operations are performed in float32.  The TDA4VM SDK currently does not support dequantize operators. With the model graph translation, we were able to convert the model parameter data types to float32 and remove the dequantize operators.  

\subsubsection{Channel-wise Padding operator }

The Mediapipe tflite models for face detection and face landmark have channel-wise padding operators. The \href{https://www.tensorflow.org/api_docs/python/tf/pad}{tf.pad} operator supports the addition of CONSTANT values using the padding attribute of the operator. The configuration parameters of the padding layer specify how much padding needs to be done along each axis of the operator’s input data. In Mediapipe models, the operator is configured to pad the input with Zero values (CONSTANT values) along the channel. However, the TDA4VM SDK does not support the specified padding operation. Therefore, the following alternative operators were experimented with in order to replace this operator without affecting layer functionality.  

\begin{enumerate}
    \item \textbf{Concatenation with zero values}: Since the channel-wise padding with Zero values adds zeros along the channel axis, the Padding operator can be replaced with a Concatenation operation. The number of zeros to be padded was determined using the padding attribute from the padding layer
    \item \textbf{Conv2D operation}: The result equivalent to zero padding operation can be achieved using a zero-padded identity matrix as a filter in the Conv2D operation. 
\end{enumerate}

Both the above-mentioned implementations can be used as alternatives for cases where channel-wise padding is not supported by the underlying hardware. However, we observe that in the case of concatenating with zeros, the zero tensors of the required size (as per the padding configuration) will have to be another input to the model which results in larger data transfer from CPU memory to the MMA memory and reduces the total inference speed of the model. Therefore, we replaced the Padding operator with Conv2D operation. Below we provide a simple example of how the padding operator can be mathematically achieved using the matrix multiplication operator which can be extended to the Conv 2D operation as well. 

For example, if the operation is expected to add a padding of 4 zeros to the output, the same can be achieved by framing a filter by concatenating an identity matrix (same dimension as input) and a matrix with zeros (padding as per requirement and dimension adapting to the identity matrix). An example of the same is represented below. Padding four zeros along the width: 






\begin{gather}
 \begin{bmatrix}
1 & 1 & 1\\
1 & 1 & 1\\
1 & 1 & 1
\end{bmatrix} 
\times
\begin{bmatrix}
1 & 0 & 0 & 0 & 0 & 0 & 0\\
0 & 1 & 0 & 0 & 0 & 0 & 0\\
0 & 0 & 1 & 0 & 0 & 0 & 0
\end{bmatrix}
=
\begin{bmatrix}
1 & 1 & 1 & 0 & 0 & 0 & 0\\
1 & 1 & 1 & 0 & 0 & 0 & 0\\
1 & 1 & 1 & 0 & 0 & 0 & 0
\end{bmatrix}
\end{gather}
\begin{flalign}
\hspace{82pt}Input\hspace{38pt}    Filter\hspace{50pt} Output && \nonumber
\end{flalign}

\subsection{Quantization }

Once the models are ported successfully to run on the MMA, the models are quantized by the TIDL compilation tool, and the inference is run on 8-bit integer precision. The C7x DSP/MMA is optimized to run integer computations (8-bit, 16-bit) and quantization of the float32 pre-trained models helps in significantly reducing the inference time as well as the bandwidth and storage.

\section{Results}
Experiments were conducted with the models mentioned in the table and we report the frames per second (FPS) results before and after doing the model graph surgery. All models are quantized to int8. 



\noindent\setlength\tabcolsep{4pt}%
\begin{tabularx}{0.8\linewidth}
{p{0.03\linewidth}p{0.17\linewidth}p{0.15\linewidth}p{0.15\linewidth}p{0.17\linewidth}p{0.2\linewidth}}
\caption{Frames per second (FPS) results before and after the model surgery}\label{tab:fps}\\
  \toprule
  Sl No & Model Name & FPS before model surgery           & FPS after model surgery                & Relative FPS improvement (\%)              & Surgery steps performed  \\ [0.5ex]
  \midrule
  1 & Mediapipe face detection model \cite{blazeface} & 121.55 & 259.84 & 113.77 \% & Padding replaced with Conv2D \\
  \addlinespace
2 & YOLOX-tiny \cite{edgeaimodelzoo} & 170 & \_ & \_ & No model surgery was performed \\
\addlinespace
3 & Mediapipe face landmark model \cite{mpfacemesh} & 163.60 & 500.97 & 206.22\% & Padding replaced with Conv2D \\
\addlinespace
4 & PFLD face landmark model \cite{pfld} & 688.59 & \_ & \_ & No model surgery \\
  \bottomrule
\end{tabularx}
\setlength{\tabcolsep}{1pt}

\subsection{DMS System level performance }

The results in Table \ref{tab:fps} indicate that the best models for face detection and face landmarks are the Mediapipe face detection model and the PFLD face landmark model. However, since the dynamic range of output tensor values for bounding box prediction $[0 - 1]$ and facial key points $[0 - \text{image width/height}]$ are drastically different, severe quantization errors were observed. Here we propose to split the output tensor into different branches for the bounding box and facial key points. However, we keep this for future work as it is beyond the scope of the model surgery experiments. Therefore, we used the YOLOX-tiny model \cite{edgeaimodelzoo} as it is already optimized and ported to the TDA4VM device which runs at ~170fps. For the face landmarks model, we chose the PFLD \cite{pfld} model which already works faster than the Mediapipe models.  

With the YOLOX-tiny model and the PFLD face landmark model, the DMS achieves 63 FPS which exceeds the target speed of 30 FPS. The detailed timing information of the DMS is given below in Table \ref{tab:timing}

Although the models after the model surgery were not used in this version of DMS, we still observe that the ported models can still enable real-time performance for the system. Additionally, the BlazeFace model can be faster than the YOLOX-tiny face detection model by fixing the quantization issue.  We are planning these as part of our ongoing roadmap of optimization on Edge AI hardware.

\noindent\setlength\tabcolsep{4pt}%
\begin{tabularx}{0.8\linewidth}
{p{0.7\linewidth}p{0.3\linewidth}}
\caption{Processing time for each perception layer in DMS. \\ \textbf{FPS = 1000/15.8 = 63.29}}\label{tab:timing}\\
  \toprule
  Processing steps & Execution time (ms) \\ [0.5ex]
  \midrule
  Input image preprocessing & 0.2 \\
  \addlinespace
  Face detection inference  & 6.1 \\
  \addlinespace
  Face detection post-process  & 0.4 \\
  \addlinespace
  Landmark image pre-processing  & 1.8 \\
  \addlinespace
  Landmark model inference  & 3.2 \\
  \addlinespace
  Landmark results post-processing  & 0.3 \\
  \addlinespace
  Blink/Yawn/Head pose computation + Visualization time  & 3.8 \\
  \hline
  Total time & 15.8 \\
  \bottomrule
\end{tabularx}
\setlength{\tabcolsep}{1pt}

\section{Conclusion}
Automated driver monitoring systems have gained a lot of attention in recent years due to government regulations. DMS implemented on edge devices are desirable due to its cost benefits. Edge AI devices, like the highly capable TI TDA4VM, are equipped with specialized hardware accelerators that help in speeding up compute-intensive DL model operations. However, getting a really high performance DMS working on these accelerator-enabled edge devices has its own challenges. In order to fully get the DL model to execute on the hardware accelerator, unsupported layers/operators will have to be replaced or removed without affecting the model functionality. We observed that once the model is fully executed on the MMA, the inference latency is reduced. With the DL models completely executed on the MMA, our DMS achieves 63 FPS which is above target speeds.

\clearpage
%
%
\bibliographystyle{splncs04}
\bibliography{egbib}
\end{document}